\documentclass[final,3p,times,twocolumn]{elsarticle}

\usepackage{amssymb}
\usepackage{amsmath}
\usepackage{graphicx}
\usepackage{booktabs}
\usepackage{booktabs}
\usepackage{multirow}
\usepackage{hyperref}
\usepackage{float}
\usepackage{placeins}
\usepackage{dblfloatfix}
\usepackage{enumitem}

\usepackage{changes}

\journal{Information Fusion}

\begin{document}

\begin{frontmatter}

\title{KID: Knowledge-Injected Dual-Head Learning for Knowledge-Grounded Harmful Meme Detection}

\author[1]{Yaocong Li}

\author[2]{Le Zhang}

\author[1]{Leihan Zhang\corref{cor1}}
\cortext[cor1]{Corresponding author}
\ead{zhangleihan@gmail.com}

\author[1]{Qiang Yan}

\affiliation[1]{organization={School of Economics and Management, Beijing University of Posts and Telecommunications},
            addressline={No. 10 Xitucheng Road, Haidian District},
            city={Beijing},
            postcode={100876},
            country={China}}

\affiliation[2]{organization={College of Computing, Beijing Information Science and Technology University},
            addressline={No. 12, Xiaoying East Road, Haidian District}, 
            city={Beijing},
            postcode={100085}, 
            country={China}}

\begin{abstract}
Internet memes have become pervasive carriers of digital culture on social platforms. However, their heavy reliance on metaphors and sociocultural context also makes them subtle vehicles for harmful content, posing significant challenges for automated content moderation. Existing approaches primarily focus on intra-modal and inter-modal signal analysis, while the understanding of implicit toxicity often depends on background knowledge that is not explicitly present in the meme itself.
To address this challenge, we propose KID, a Knowledge-Injected Dual-Head Learning framework for knowledge-grounded harmful meme detection. KID adopts a label-constrained distillation paradigm to decompose complex meme understanding into structured reasoning chains that explicitly link visual evidence, background knowledge, and classification labels. These chains guide the learning process by grounding external knowledge in meme-specific contexts. In addition, KID employs a dual-head architecture that jointly optimizes semantic generation and classification objectives, enabling aligned linguistic reasoning while maintaining stable decision boundaries.
Extensive experiments on five multilingual datasets spanning English, Chinese, and low-resource Bengali demonstrate that KID achieves SOTA performance on both binary and multi-label harmful meme detection tasks, improving over previous best methods by 2.1\%--19.7\% across primary evaluation metrics. Ablation studies further confirm the effectiveness of knowledge injection and dual-head joint learning, highlighting their complementary contributions to robust and generalizable meme understanding. The code and data are available at \url{https://github.com/PotatoDog1669/KID}.
\end{abstract}


\end{frontmatter}


\section{Introduction}
\label{sec:intro}

Internet memes are core carriers of digital culture. They spread rapidly across social platforms through their unique coupling of visual and textual modalities. While often serving as vehicles for humor and cultural expression, they have increasingly become hidden channels for harmful content such as hate speech, stereotypes, and gender discrimination \cite{kiela2020hateful}. The toxicity of such harmful memes rarely appears explicitly within a single modality. Instead, it is hidden within the complex interplay between visual symbols and textual information, heavily relying on metaphorical rhetoric, cultural background, and cross-modal semantic interactions \cite{pramanick2021detecting}. For instance, memes can strategically pair female imagery with objectifying symbols to convey misogyny, or leverage regional cultural symbols to target specific communities---such content not only degrades the online ecosystem but may also worsen social tensions, presenting significant challenges for content governance \cite{fersini2022semeval}.

\FloatBarrier
\begin{figure*}[!t]
\centering
\includegraphics[width=0.95\textwidth]{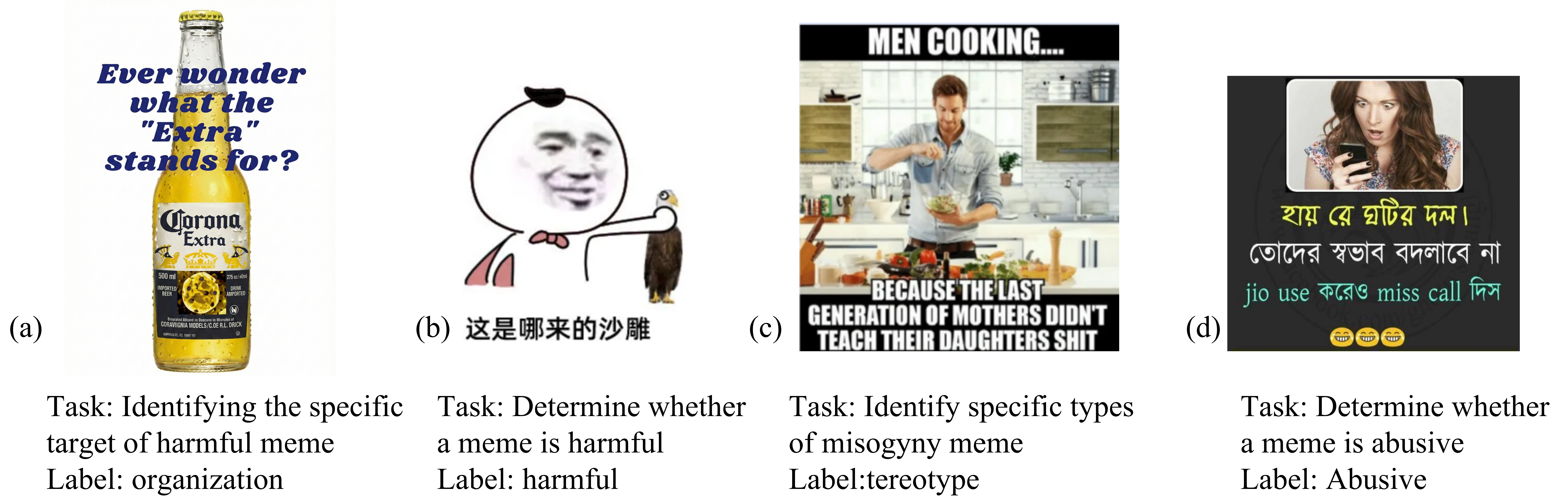}
\caption{Representative Examples of Harmful Memes: (a) English: brand-pandemic association, (b) Chinese: cultural slang (translation: ``Where does this sand sculpture come from?''), (c) English: gender stereotype, and (d) Bengali: abusive content (translation: ``Oh dear, you lot. Your habits will never change. Even after using Jio, you still give a missed call.'').}
\label{fig:examples}
\end{figure*}

Detecting toxic content in memes has attracted growing research interest. Existing approaches largely rely on unimodal feature extraction with simple cross-modal fusion, or leverage pre-trained vision language models to capture explicit semantic correlations between images and text \cite{early1,early2,visualbert,clip}. More recently, multimodal large language models (MLLMs) have been explored due to their strong vision–language alignment and extensive world knowledge \cite{lmm1,lmm2}. Despite these advances, current methods remain limited in understanding the implicit and culturally grounded semantics of memes. Toxic intent often emerges from metaphorical associations, symbolic imagery, and background knowledge that cannot be inferred from surface-level visual textual alignment alone. As a result, existing approaches struggle with cross-scenario generalization, multi-label attribute recognition, and low-resource language settings \cite{kermit,cao2022prompting,lmm3,lmm4}.

To better illustrate these bottlenecks, Figure~\ref{fig:examples} presents four representative harmful meme examples across different languages, highlighting the core challenges of this task.
In example (a) from the HarMeme dataset, the image shows an ordinary Corona beer, while the accompanying text implicitly exploits the phonetic similarity between “Corona” and “Coronavirus,” evoking associations with the COVID-19 pandemic. Correct interpretation requires recognizing that the meme targets the Corona \textbf{organization} rather than society in general, revealing the difficulty of accurately identifying attack targets. The harmful intent only becomes clear when \textbf{visual symbols (brand name)}, \textbf{background knowledge (the Corona–Coronavirus association)}, and the \textbf{attack target (organization)} are jointly considered.
Example (b) from the ToxiCN\_MM dataset highlights the importance of cultural context. The Chinese phrase “sha diao” (literally “sand sculpture”) is a homophonic internet slang commonly used as a vulgar insult. When combined with a pointing gesture, the meme shifts the phrase from possible self-mockery to a direct personal attack, showing that semantic ambiguity can only be resolved with language- and culture-specific knowledge.
In example (c) from the MAMI dataset, harmful meaning arises from implicit gender stereotypes, conveyed by pairing a cooking scene with text suggesting women’s failure in traditional roles. Example (d) from the BanglaAbuseMeme dataset further shows that similar challenges exist in low-resource languages, where abusive intent often relies on culturally specific expressions.
Overall, these examples reveal a shared challenge: toxic meaning in memes depends on effectively linking background knowledge with visual and textual cues. However, existing methods lack explicit mechanisms to establish this link, leading to failures in correctly identifying toxic intent. We refer to this challenge as the \textbf{knowledge-context disconnect} problem.

To address the knowledge–context disconnect problem, we identify two fundamental limitations in existing approaches. First, although MLLMs acquire rich background knowledge during pre-training, they lack explicit mechanisms to bind this knowledge to concrete visual symbols and textual contexts in memes, resulting in weak grounding when inferring toxic intent. Second, current research predominantly focuses on binary classification, with relatively limited attention to multi-label attribute recognition and low-resource language scenarios.

Motivated by these limitations, we propose a \textbf{Knowledge-Injected Dual-Head Learning Framework}, abbreviated as \textbf{KID}, for harmful meme detection. The core idea of KID is to explicitly guide models to ground background knowledge in meme-specific visual–textual contexts while maintaining stable and reliable decision boundaries across diverse tasks. To this end, KID incorporates a structured knowledge injection strategy that aligns background knowledge with multimodal evidence and classification objectives, together with a dual-head learning architecture that jointly optimizes semantic understanding and decision stability. By regulating the amount of injected knowledge, KID further balances information gain and noise suppression, enabling large models to effectively utilize their knowledge rather than rely on superficial pattern memorization.

The main contributions of this work are summarized as follows:
\begin{enumerate}[label=\roman*]
    \item We propose an \textbf{entity-anchored knowledge injection mechanism} that explicitly binds background knowledge to visual entities through structured reasoning chains, effectively addressing the \textbf{knowledge-context disconnect} in harmful meme understanding.
    
    \item We introduce \textbf{KID}, a \textbf{knowledge-injected dual-head learning framework} that jointly optimizes label generation and classification stability through multi-task learning, improving both task alignment and robustness.
    
    \item Extensive experiments on five multilingual datasets (English, Chinese, and Bengali) demonstrate that KID achieves \textbf{SOTA performance}, strong cross-lingual generalization, and effective control of knowledge injection.
\end{enumerate}

\section{Related Work}
\label{sec:related}

\subsection{Multimodal Modeling for Harmful Meme Detection}

Existing approaches for harmful meme detection can be broadly categorized into three lines of research. Overall, prior work has made steady progress in modeling visual–textual correlations, yet remains limited in capturing the implicit and culturally grounded semantics that characterize harmful memes.

The first line of work focuses on unimodal feature extraction followed by basic cross-modal fusion. Visual encoders such as ResNet \cite{resnet} and ViT \cite{vit} are commonly used to extract image representations, while textual semantics are modeled using language models like BERT \cite{bert} and RoBERTa \cite{roberta}. These features are aligned through concatenation, weighted fusion, or cross-attention mechanisms \cite{crossattn} to obtain unified multimodal representations. While effective for capturing explicit visual–textual correlations, such methods largely operate at the surface semantic level and struggle to recognize harmful expressions that rely on cultural background or metaphorical reasoning.

The second line of research leverages contrastive vision–language pre-training models, most notably CLIP \cite{clip}, to enhance multimodal alignment. Building upon CLIP’s global vision–text representations, various task-specific adaptations have been proposed, including prompt-based methods \cite{procap}, feature interaction modeling \cite{hateclipper}, meme-oriented architectural designs \cite{memeclip}, and retrieval-guided contrastive learning strategies \cite{rgcl}. Although these approaches improve fine-grained cross-modal interaction and generalization, they still rely primarily on implicit alignment signals and lack explicit mechanisms for grounding background knowledge in meme-specific contexts.

More recently, a third category explores MLLMs, which integrate discriminative modeling with multimodal generative capabilities \cite{lmm_shift1,lmm_shift2}. Decoder-based MLLMs enable natural language reasoning and explanation, offering advantages in interpretability \cite{lmm2}. While supervised fine-tuning has shown promising results for hateful meme detection \cite{lmm_shift1}, MLLM-based solutions remain underexplored in this domain. Moreover, prior studies indicate that carefully fine-tuned CLIP-based models can rival or even outperform larger MLLMs on certain tasks \cite{rgcl}, suggesting that model scaling alone does not resolve the complex reasoning challenges in meme understanding.

Despite these advances, a shared limitation persists across all three categories. Although MLLMs and vision-language models acquire rich general knowledge during pre-training, they often fail to effectively ground this knowledge in specific cultural allusions or metaphorical scenes within memes. When faced with implicit reasoning chains such as \emph{visual symbol → background knowledge → harmful intent}, models frequently misinterpret or overlook toxicity due to a disconnect between general knowledge and concrete visual-textual contexts. This challenge corresponds to the previously defined \textbf{knowledge-context disconnect} problem.

\subsection{Knowledge Enhancement in Harmful Content Detection}

Understanding harmful memes often requires implicit sociocultural background and commonsense reasoning, which cannot be adequately captured through surface-level vision-text alignment alone. As a result, recent studies have explored incorporating external knowledge into multimodal frameworks to enhance models’ ability to interpret metaphors, cultural allusions, and implicit intent. Overall, existing knowledge enhancement approaches can be broadly categorized into \emph{structured explicit enhancement} and \emph{unstructured implicit enhancement}.  

\textbf{Structured explicit enhancement} methods leverage external knowledge bases to construct explicit semantic associations. Representative works include KERMIT, which maps entities detected in memes to commonsense knowledge graphs such as ConceptNet and employs memory-augmented networks to filter relevant nodes \cite{kermit}, as well as KnowMeme, which builds entity-centric graph structures to assist classification decisions \cite{knowmeme}. While these approaches provide clear and interpretable pathways for knowledge injection, they are constrained by the limited coverage and delayed updates of static knowledge graphs. Consequently, they often struggle to adapt to the fast-evolving nature of internet memes and to tightly bind injected knowledge to task-specific classification objectives.  

\textbf{Unstructured implicit enhancement} methods adopt lighter-weight strategies to activate background knowledge without explicit graph construction. Common practices include prompt-based techniques that guide large models to recall relevant commonsense knowledge \cite{prompt_knowledge}, as well as retrieval-augmented generation (RAG) approaches that concatenate externally retrieved text with meme content as model input \cite{rag}. Although these methods are easy to integrate with pre-trained models, the lack of effective filtering and control mechanisms often introduces substantial noise. Irrelevant or semantically drifted information may interfere with intent recognition and even lead to knowledge hallucination.  

Despite their differences, both categories share a fundamental limitation: the difficulty of \textbf{aligning external knowledge with meme-specific contexts without semantic drift}. Existing approaches lack explicit and controllable mechanisms to anchor external knowledge to key visual symbols in memes and to form stable \emph{evidence--knowledge--label} reasoning chains. Moreover, coarse-grained knowledge injection frequently results in negative transfer, where introducing more knowledge degrades performance. These issues substantially limit model robustness when handling implicit toxicity, multi-label discrimination, and cross-cultural meme understanding.

\begin{figure*}[!htbp]
\centering
\includegraphics[width=1.0\textwidth]{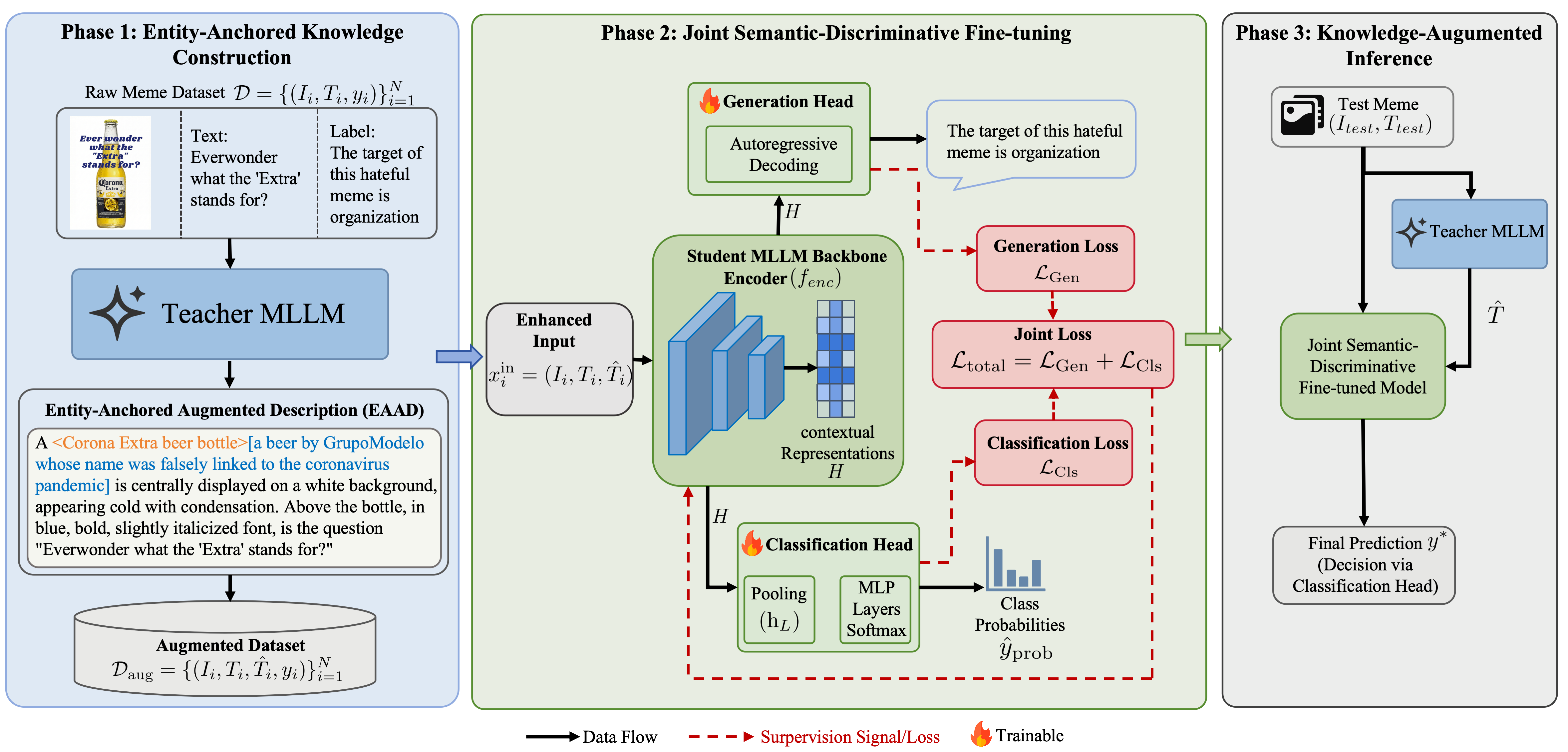}
\caption{The framework of KID.}
\label{fig:framework}
\end{figure*}

\section{Methodology}
\label{sec:method}

\subsection{Preliminaries}

\textbf{Problem Formulation.} We formalize harmful meme detection as a multimodal classification problem that relies on deep semantic reasoning. Let $\mathcal{D} = \{(x_i, y_i)\}_{i=1}^{|\mathcal{D}|}$ denote a dataset containing $|\mathcal{D}|$ samples, where each input sample $x_i = (I_i, T_i)$ consists of a visual modality $I_i$ (image) and a textual modality $T_i$ (embedded OCR text or caption). Our goal is to learn a mapping function $\mathcal{F}_{\Theta}: \mathcal{X} \to \mathcal{Y}$ to accurately predict the target label $y_i$ from the predefined label space $\mathcal{Y}$.

\textbf{Backbone Encoder and Classification Paradigm.} To handle complex cross-modal interactions, we adopt a pre-trained multimodal large language model (MLLM) as the backbone feature extractor, denoted as $f_{\text{enc}}$. Given a multimodal input pair, the encoder maps it to a sequence of high-dimensional contextual representations:
\begin{equation}
\mathbf{H} = f_{\text{enc}}(I, T; \Theta_{\text{enc}}),
\end{equation}
where $\mathbf{H} \in \mathbb{R}^{L \times d}$ represents the hidden state sequence output from the last Transformer layer, $L$ is the sequence length, and $d$ is the hidden dimension.

In conventional classification paradigms, models typically apply a linear classification head directly on the pooled representation $\mathbf{h}_{\text{pool}}$ to compute class probabilities:
\begin{equation}
P(y \mid I, T) = \text{Softmax}(\mathbf{W}_c \mathbf{h}_{\text{pool}} + \mathbf{b}_c).
\end{equation}

However, this end-to-end paradigm often struggles to capture the cultural metaphors and background knowledge implicit in memes, causing models to overfit to superficial visual-textual statistical correlations. Therefore, our method aims to reconstruct this process through explicit knowledge enhancement.

\subsection{Overview of the Proposed Framework}

To address the knowledge-context disconnect problem, we propose a \textbf{Knowledge-Injected Dual-Head Learning (KID)} framework. Rather than directly mapping from raw pixels to labels, KID explicitly models the evidence-knowledge-label reasoning chain. As shown in Figure~\ref{fig:framework}, our method comprises three progressive phases:

\begin{enumerate}[label=\roman*]
    \item \textbf{Entity-Anchored Knowledge Construction.} We leverage the open-world knowledge capabilities of multimodal large models to explicitly transform the implicit background information in memes into text. Through an ``entity-anchoring'' strategy, we precisely bind abstract knowledge to concrete visual entities, constructing a knowledge-enhanced dataset.
    
    \item \textbf{Joint Semantic-Discriminative Learning.} We design a dual-head architecture that simultaneously learns ``what it is'' (semantic alignment) and ``which class it belongs to'' (decision) on the enhanced input. This multi-task learning forces the model to align its internal representation with both semantic concepts and decision boundaries.
    
    \item \textbf{Knowledge-Augmented Inference.} During inference, we adopt a \textbf{Test-Time Knowledge Injection} strategy. We first utilize a multimodal large model to generate knowledge context on-the-fly, then use the fine-tuned dual-head model for final prediction, thereby ensuring knowledge consistency between inference and training phases.
\end{enumerate}

\subsection{Entity-Anchored Knowledge Construction}

\textbf{Motivation.} Most student models used for fine-tuning have limited parameters and lack the extensive popular culture, political metaphor, and hate symbol knowledge required to understand memes. To address this ``knowledge deficiency'' problem, we need to transform implicit background knowledge into explicit textual context readable by the model. However, simply generating a generic description often introduces noise or hallucinations. Therefore, we propose an \textbf{Entity-Anchored} strategy to ensure that injected knowledge tightly revolves around key visual elements in the meme.

\textbf{Structured Knowledge Generation.} For each sample $(I_i, T_i)$ in the dataset, we utilize an advanced multimodal large model (Teacher MLLM), denoted as $\mathcal{T}$, as the knowledge engine to generate an Interleaved Augmented Description $\hat{T}_i$:
\begin{equation}
\hat{T}_i = \mathcal{T}(I_i, T_i; \Theta_{\text{teacher}})
\end{equation}

This description follows a strict structure: the model first identifies objective entities in the image, then immediately inserts the background knowledge of that entity after it. Formally, the generated augmented text sequence is:
\begin{equation}
\hat{T}_i = \{ w_1, \dots, \langle \text{Entity}_j \rangle, [\text{Knowledge}_j], \dots, w_M \}
\end{equation}

For instance, for a meme containing ``Pepe the Frog,'' the generated sequence might be: ``The image shows $\langle$Pepe the Frog$\rangle$ [an internet meme symbol often used by far-right groups], looking at...''

This structured $\langle$Entity$\rangle$-[Knowledge] pairing mechanism effectively establishes strong associations between visual signals and abstract semantics. Through this process, we transform the original dataset $\mathcal{D}$ into a knowledge-augmented dataset $\mathcal{D}_{\text{aug}} = \{(I_i, T_i, \hat{T}_i, y_i)\}_{i=1}^{|\mathcal{D}|}$.

\begin{table*}[t]
\centering
\caption{Dataset Statistics and Evaluation Metrics.}
\label{tab:datasets}
\begin{tabular}{lllclll}
\toprule
Task & Language & Dataset & Classes & Train & Val & Test \\
\midrule
$\text{B}$ & English & Hateful Memes & Harmful / Non-Harmful & 8,500 & 500 & 1,000 \\
$\text{M}$ & English & HarMeme & Individual / Organization / Community / Society & 2,514 & 61 & 124 \\
$\text{B}$ & English & MAMI Task A & Misogynous / Non-misogynous & 9,000 & 1,000 & 1,000 \\
$\text{M}$ & English & MAMI Task B & Shaming / Stereotype / Objectification / Violence & 9,000 & 1,000 & 1,000 \\
$\text{B}$ & Chinese & ToxiCN\_MM Task A & Harmful / Non-Harmful & 7,200 & 2,400 & 2,400 \\
$\text{M}'$ & Chinese & ToxiCN\_MM Task B & Targeted / Sexual / Offense / Dispirited & 2,333 & 741 & 741 \\
$\text{B}$ & Bengali & BanglaAbuseMeme A & Abusive / Non-Abusive & 3,233 & 405 & 405 \\
$\text{M}$ & Bengali & BanglaAbuseMeme B & Individual / Social / Gender / Political / etc. & 2,270 & 285 & 285 \\
\bottomrule
\end{tabular}
\par\smallskip\footnotesize\raggedright
\textit{Note:} ``$\text{B}$'' refers to binary classification; ``$\text{M}$'' refers to single-label multi-class classification; ``$\text{M}'$'' refers to multi-label multi-class classification.
\end{table*}

\subsection{Joint Semantic-Discriminative Learning}

\textbf{Input Representation.} During fine-tuning, we concatenate the original modalities with the constructed knowledge context to form the enhanced input $x_i^{\text{in}}$:
\begin{equation}
x_i^{\text{in}} = (I_i, T_i, \hat{T}_i)
\end{equation}

\textbf{Dual-Head Architecture Design.} To leverage the complementary strengths of generative and discriminative paradigms, we design parallel dual-head structures at the output end $\mathbf{H}$ of the backbone network:

\textbf{(1) Semantic Generation Head:}

This head serves as an auxiliary task that encourages the model to map its internal representations to the natural language space. We retain the causal language modeling (CLM) head, requiring the model to output the classification label in natural language form.

Specifically, given the enhanced input $x_i^{\text{in}}$, the model needs to autoregressively generate the target sequence $S_{\text{text}}$ (e.g., ``The target is Individual''). The prediction probability for the $t$-th token is computed as:
\begin{equation}
P(w_t \mid x_i^{\text{in}}, w_{<t}) = \text{Softmax}(\mathbf{H}_t \mathbf{W}_{\text{gen}} + \mathbf{b}_{\text{gen}})
\end{equation}
where $\mathbf{W}_{\text{gen}}$ projects the hidden state to the vocabulary space. This auxiliary task promotes consistency between the learned features and the linguistic structure of the classification task.

\textbf{(2) Discriminative Classification Head:}

To obtain stable decision boundaries, we introduce a dedicated classification head.

We directly use the hidden state at the final token position $\mathbf{h}_L$ from the enhanced sequence features $\mathbf{H}$ as the classification feature (i.e., \textbf{last-token pooling}), where $\mathbf{H} = [\mathbf{h}_1, \dots, \mathbf{h}_L]$. We then map this vector to class logits through a Multi-Layer Perceptron (MLP):
\begin{equation}
\hat{y}_{\text{prob}} = \text{Softmax}\left( \sigma(\mathbf{h}_L\mathbf{W}_1 + \mathbf{b}_1)\mathbf{W}_2 + \mathbf{b}_2 \right)
\end{equation}

This head focuses on optimizing decision boundaries, ensuring the model's discriminative capability within the closed label space $\mathcal{Y}$.

\textbf{Multi-Task Joint Optimization.} Model training adopts an end-to-end joint optimization strategy. The total loss function $\mathcal{L}_{\text{total}}$ is the sum of the two task-specific losses:
\begin{equation}
\mathcal{L}_{\text{total}} = \mathcal{L}_{\text{Gen}} + \mathcal{L}_{\text{Cls}}
\end{equation}

Specifically, the semantic generation loss $\mathcal{L}_{\text{Gen}}$ is the standard negative log-likelihood over the target token sequence:
\begin{equation}
\mathcal{L}_{\text{Gen}} = -\frac{1}{|S|}\sum_{t=1}^{|S|} \log P(w_t \mid x^{\text{in}}, w_{<t})
\end{equation}
where $|S|$ denotes the length of the target sequence $S_{\text{text}}$.

The classification loss $\mathcal{L}_{\text{Cls}}$ is the standard cross-entropy loss:
\begin{equation}
\mathcal{L}_{\text{Cls}} = -\sum_{c \in \mathcal{Y}} y_c \log \hat{y}_{\text{prob},c}
\end{equation}
where $y_c$ is the ground-truth one-hot label and $\hat{y}_{\text{prob},c}$ is the predicted probability for class $c$.

\subsection{Knowledge-Augmented Inference}

In practical application scenarios, test samples are often novel and do not contain pre-annotated knowledge. To maintain feature distribution consistency with the training phase and leverage the advantages of knowledge enhancement, we adopt a \textbf{Test-Time Knowledge Injection} strategy. The inference process is divided into two strict steps:

\textbf{On-the-fly Knowledge Context Generation.}

For an input test sample $(I_i, T_i)$, we first utilize a general multimodal large model (comparable in capability to the model in Phase 1) to generate the augmented description $\hat{T}_i$ in real-time.

This process is equivalent to dynamically mounting an external knowledge base at inference time, explicitly parsing the implicit entities in the meme (such as specific political figures or subculture symbols) into textual descriptions. This step does not involve classification decisions; it is solely responsible for information enrichment.

\textbf{Dual-Head Prediction.}

The generated augmented description $\hat{T}_i$ is fed into the fine-tuned dual-head model together with the original input. The model simultaneously produces two forms of output:

\begin{enumerate}[label=\roman*]
    \item \textbf{Semantic Label Output}: The semantic generation head produces the classification label in natural language form to ensure alignment with the linguistic task structure. We utilize dataset-specific templates:
    \begin{itemize}
        \item \textbf{Hateful Memes}/ \textbf{ToxiCN\_MM Task A}: ``This meme is harmful / non-harmful''
        \item \textbf{BanglaAbuseMeme Task A}: ``This meme is abusive / non-abusive''
        \item \textbf{MAMI Task A}: ``Yes / No, this meme is (not) misogynous''
        \item \textbf{HarMeme}/ \textbf{BanglaAbuseMeme Task B}: ``The target of this hateful meme is [Target]''
        \item \textbf{ToxiCN\_MM Task B}: ``This meme belongs to the category: [Category]''
        \item \textbf{MAMI Task B}: ``Categories: [Category1, Category2]''
    \end{itemize}
    \item \textbf{Probabilistic Quantitative Output}: The classification head outputs precise class confidence scores, supporting binary, single-label multi-class, and multi-label multi-class classification scenarios.
\end{enumerate}

The final classification decision $y^*$ is determined by the classification head to ensure superior performance on quantitative metrics such as AUC:
\begin{equation}
y^* = \operatorname*{argmax}_{c \in \mathcal{Y}} \left[ \hat{y}_{\text{prob}} \right]_c
\end{equation}

\section{Experiments}
\label{sec:experiments}

\subsection{Datasets}

To comprehensively evaluate the performance and generalizability of our proposed framework, we conduct experiments on five multimodal datasets covering different languages (English, Chinese, and Bengali) and varying granularities of toxicity. Depending on the task formulation, we employ \textbf{Area Under the Curve (AUC)} and \textbf{Accuracy (ACC)} for binary classification tasks, while \textbf{Macro-F1} and \textbf{ACC} are used for fine-grained multi-class tasks. Table~\ref{tab:datasets} summarizes the detailed statistics and corresponding metrics.

\begin{table*}[t]
\centering
\caption{Performance on Multi-class Classification Tasks (Macro-F1 / ACC).}
\label{tab:multiclass}
\begin{tabular*}{\textwidth}{@{\extracolsep{\fill}}lcccc}
\toprule
Model & HarMeme & MAMI Task B & ToxiCN\_MM B & BanglaAbuse B \\
\midrule
\textbf{Previous SOTA} & 69.65/77.95 \cite{pramanick2021detecting} & 73.00/--- \cite{srcb} & 62.17/--- \cite{toxicn} & --- \\
 \midrule
CLIP & 62.14/72.47 & --- & 57.85/--- & --- \\
MOMENTA & 69.65/77.95 & --- & --- & --- \\
Gemini-2.5-Flash (Zero) & 35.41/45.97 & 73.44/73.70 & 40.63/38.06 & 49.40/58.60 \\
Gemini-2.5-Flash (Few) & 53.56/67.32 & 75.34/74.22 & 56.32/63.72 & 55.24/63.15 \\
Qwen2.5-VL-7B (SFT) & 60.84/75.27 & 70.35/78.32 & 74.23/79.16 & 69.88/74.26 \\
Qwen2.5-VL-7B (Dual-Head) & 63.40/79.00 & 75.36/83.56 & 78.29/81.38 & 74.89/77.28 \\
Qwen2.5-VL-7B (Dual-Head + Caption) & 63.70/79.20 & 74.54/82.86 & 78.76/80.67 & 74.02/78.85 \\
\multirow{2}{*}{\textbf{KID (Ours)}} & 65.50/\textbf{80.70} & \textbf{79.60/87.50} & \textbf{81.84/86.42} & \textbf{76.82/80.18} \\
 & \textit{(+2.75)} & \textit{(+6.60/---)} & \textit{(+19.67/---)} & \textit{(---)} \\
\bottomrule
\end{tabular*}
{\par\smallskip\footnotesize\raggedright
\textit{Note:} ``---'' indicates: for Previous SOTA, the original work did not report this metric; for other baselines, no experiments were conducted on this dataset.
\par}

\vspace{5mm}

\caption{Performance on Binary Classification Tasks.}
\label{tab:binary}
\begin{tabular*}{\textwidth}{@{\extracolsep{\fill}}lcccc}
\toprule
Model & Hateful Memes & MAMI Task A & ToxiCN\_MM A & BanglaAbuse A \\
\midrule
\textbf{Previous SOTA} & 91.10/82.10 \cite{lmmrgcl} & 91.20/79.70 \cite{lmmrgcl} & 80.33/--- \cite{alfus} & 70.50/71.80 \cite{clip} \\
 \midrule
CLIP & 79.80/72.00 & 77.70/68.40 & 79.54/--- & 70.50/71.80 \\
MOMENTA & 69.20/61.30 & 81.70/72.10 & --- & --- \\
HateCLIPper & 85.50/76.10 & 87.20/74.80 & --- & --- \\

Gemini-2.5-Flash (Zero) & 74.26/70.35 & 63.30/60.56 & 59.99/74.29 & 75.75/77.04 \\
Gemini-2.5-Flash (Few) & 88.37/85.27 & 83.77/78.43 & 73.63/80.34 & 80.36/85.37 \\
Qwen2.5-VL-7B (SFT) & 86.30/78.60 & 82.60/72.40 & 76.37/78.23 & 68.32/70.37 \\
Qwen2.5-VL-7B (Dual-Head) & 91.10/83.90 & 92.00/82.00 & 81.86/84.75 & 70.13/75.09 \\
Qwen2.5-VL-7B (Dual-Head + Caption) & 90.80/83.40 & 91.19/79.65 & 81.56/84.55 & 71.11/76.12 \\
\multirow{2}{*}{\textbf{KID (Ours)}} & \textbf{93.24/85.35} & \textbf{97.20/89.30} & \textbf{84.46/87.33} & \textbf{89.94/90.31} \\
 & \textit{(+2.14/+3.25)} & \textit{(+6.00/+9.60)} & \textit{(+4.13/---)} & \textit{(+19.44/+18.51)} \\
\bottomrule
\end{tabular*}
{\par\smallskip\footnotesize\raggedright
\textit{Note:} We follow the original benchmark protocols---Hateful Memes and MAMI report AUC/ACC, while ToxiCN\_MM and BanglaAbuseMeme report F1/ACC as per their official evaluation metrics. ``---'' indicates: for Previous SOTA, the original work did not report this metric; for other baselines, no experiments were conducted on this dataset.
\par}
\end{table*}

\textbf{Hateful Memes} \cite{kiela2020hateful} is a widely used benchmark dataset for multimodal hate speech detection, constructed by Facebook AI. It comprises 10,000 samples designed to be ``benign confounders''---where neither the image nor text alone is hateful, but their combination conveys harmful intent.

\textbf{HarMeme} \cite{pramanick2021detecting} focuses on identifying the specific targets of toxicity in COVID-19 related memes. It categorizes harmful memes into four classes: \textit{Individual}, \textit{Organization}, \textit{Community}, and \textit{Society}, enabling fine-grained target analysis.

\textbf{MAMI} \cite{fersini2022semeval} (Multimedia Automatic Misogyny Identification) contains two sub-tasks. \textbf{Task A} is a binary classification task designed to detect misogynous content. \textbf{Task B} further involves multi-label classification to identify specific types of misogyny including \textit{Shaming}, \textit{Stereotype}, \textit{Objectification}, and \textit{Violence}.

\textbf{ToxiCN\_MM} \cite{toxicn} is a large-scale Chinese toxic meme dataset. \textbf{Task A} focuses on binary toxic detection, while \textbf{Task B} addresses fine-grained toxicity classification.

\textbf{BanglaAbuseMeme} \cite{bangla} is employed to assess cross-lingual robustness on low-resource languages. \textbf{Task A} performs binary abusive detection, while \textbf{Task B} focuses on multi-class target identification across seven categories.

\subsection{Experimental Setup}

\textbf{Implementation Details.} We adopt Qwen2.5-VL-7B \cite{qwen} as the backbone multimodal language model for fine-tuning. For knowledge augmentation, we utilize Gemini-2.5-Flash \cite{gemini} as the teacher model to generate entity-anchored knowledge descriptions. All experiments are conducted on NVIDIA RTX A6000 GPUs.

We employ QLoRA \cite{qlora} with INT4 quantization for parameter-efficient fine-tuning. The training configuration is as follows: batch size of 16, learning rate of 2.0e-5 for the LLM backbone and 1.0e-4 for the MLP classification head, maximum of 4 training epochs. For regularization, we apply input dropout of 0.1 and hidden dropout of 0.2.

\textbf{Baselines.} We compare our method against a diverse set of baselines spanning traditional multimodal fusion approaches, contrastive learning methods, and large multimodal models. For CLIP-based methods, we include the standard CLIP \cite{clip} with a linear classification head, MOMENTA \cite{pramanick2021detecting} which models entity-level interactions, and HateCLIPper \cite{hateclipper} that employs Feature Interaction Matrices for intermediate fusion. We also compare with retrieval-augmented approaches including the LMM-enhanced variant LMM-RGCL \cite{lmmrgcl}. For dataset-specific comparisons, we include SRCB \cite{srcb} (the winning solution of the MAMI challenge), ALFUS \cite{alfus} (designed for Chinese toxic meme detection), and Fusion-MKE \cite{toxicn} (a multi-knowledge enhanced framework). Finally, to evaluate the role of fine-tuning, we compare with Gemini-2.5-Flash under zero-shot and few-shot settings, as well as standard supervised fine-tuning of Qwen2.5-VL-7B without our dual-head architecture.

\subsection{Main Results}

\subsubsection{Results on Multi-class Classification Tasks}

Table~\ref{tab:multiclass} presents the performance comparison on multi-class classification tasks.

Our KID framework achieves consistent improvements across all four multi-class datasets. On MAMI Task B (multi-label misogyny classification), KID achieves 79.60\% Macro-F1, surpassing the previous best of 73.00\%. The most significant improvements are observed on ToxiCN\_MM Task B (+19.67\% over Fusion-MKE) and BanglaAbuseMeme Task B, demonstrating the framework's effectiveness on both Chinese and low-resource Bengali languages.

\subsubsection{Results on Binary Classification Tasks}

Table~\ref{tab:binary} presents the performance comparison on binary classification tasks.

On binary classification tasks, KID demonstrates even more pronounced improvements. On the challenging Hateful Memes benchmark, we achieve 93.24\% AUC, surpassing the previous best LMM-RGCL (91.10\%) by 2.14 percentage points. The most striking improvement is observed on MAMI Task A, where KID achieves 97.20\% AUC---a 6.0 percentage point improvement over LMM-RGCL. On BanglaAbuseMeme Task A, KID achieves 89.94\% Macro-F1, representing a remarkable 19.44 percentage point improvement over the previous best.

\textbf{Cross-lingual Transfer Analysis.} The remarkable performance on BanglaAbuseMeme deserves special attention. We hypothesize that the entity-anchored knowledge effectively bridges linguistic gaps by providing explicit cultural context in the target language. This finding suggests that structured knowledge injection can serve as an effective mechanism for cross-lingual transfer in multimodal harmful content detection.

\textbf{Statistical Significance.} To validate the robustness of our results, we conduct paired t-tests comparing KID against the strongest baseline (Qwen2.5-VL-7B Dual-Head) across three independent runs with different random seeds. The improvements are statistically significant on all datasets ($p < 0.05$). Specifically, on Hateful Memes, we observe $p = 0.012$; on MAMI Task A, $p = 0.003$; on ToxiCN\_MM Task A, $p = 0.021$; and on BanglaAbuseMeme Task A, $p = 0.008$. These results confirm that the performance gains of our knowledge injection strategy are not attributable to random variance.

\subsection{Ablation Studies}

\subsubsection{Effect of Knowledge Injection Quantity}

\begin{table}[t]
\centering
\caption{Effect of Knowledge Injection Quantity on Multi-class Tasks.}
\label{tab:ablation_multiclass}
\resizebox{\columnwidth}{!}{
\begin{tabular}{ccccc}
\toprule
N & HarMeme & MAMI B & ToxiCN B & Bangla B \\
\midrule
0 & 54.46/66.94 & 74.54/82.86 & 78.76/80.67 & 74.02/78.85 \\
1 & 57.34/73.39 & 78.53/86.46 & 80.73/84.19 & 75.51/79.53 \\
2 & \textbf{60.41/76.61} & \textbf{79.60/87.50} & \textbf{81.84/84.62} & \textbf{76.82/80.18} \\
3 & 55.55/70.97 & 79.21/86.26 & 80.29/84.38 & 76.13/80.18 \\
4 & 57.59/74.19 & 78.97/86.76 & 79.51/82.07 & 75.67/77.88 \\
5 & 61.37/75.00 & 77.10/85.28 & 79.41/82.46 & 73.61/78.47 \\
\bottomrule
\end{tabular}
}
\par\smallskip\footnotesize\raggedright
\textit{Note:} All metrics reported as Macro-F1 / ACC.
\vspace{5mm}
\caption{Effect of Knowledge Injection Quantity on Binary Tasks.}
\label{tab:ablation_binary}
\resizebox{\columnwidth}{!}{
\begin{tabular}{ccccc}
\toprule
N & Hateful & MAMI A & ToxiCN A & Bangla A \\
\midrule
0 & 90.80/83.40 & 91.19/79.65 & 82.55/85.67 & 71.11/76.12 \\
1 & 92.63/84.26 & 95.80/89.18 & \textbf{84.46/87.33} & 89.64/90.17 \\
2 & \textbf{93.24/85.35} & 96.43/88.69 & 83.46/86.33 & \textbf{89.94/90.31} \\
3 & 92.39/85.03 & \textbf{97.20/89.30} & 82.55/85.58 & 88.93/89.95 \\
4 & 91.74/84.73 & 96.80/88.35 & 82.84/85.75 & 88.36/88.93 \\
5 & 90.93/83.62 & 96.39/89.62 & 84.07/86.83 & 88.39/89.72 \\
\bottomrule
\end{tabular}
}
\par\smallskip\footnotesize\raggedright
\textit{Note:} Hateful Memes and MAMI report AUC/ACC; ToxiCN and Bangla report F1/ACC.
\end{table}

\begin{figure}[t]
\centering
\includegraphics[width=1\columnwidth]{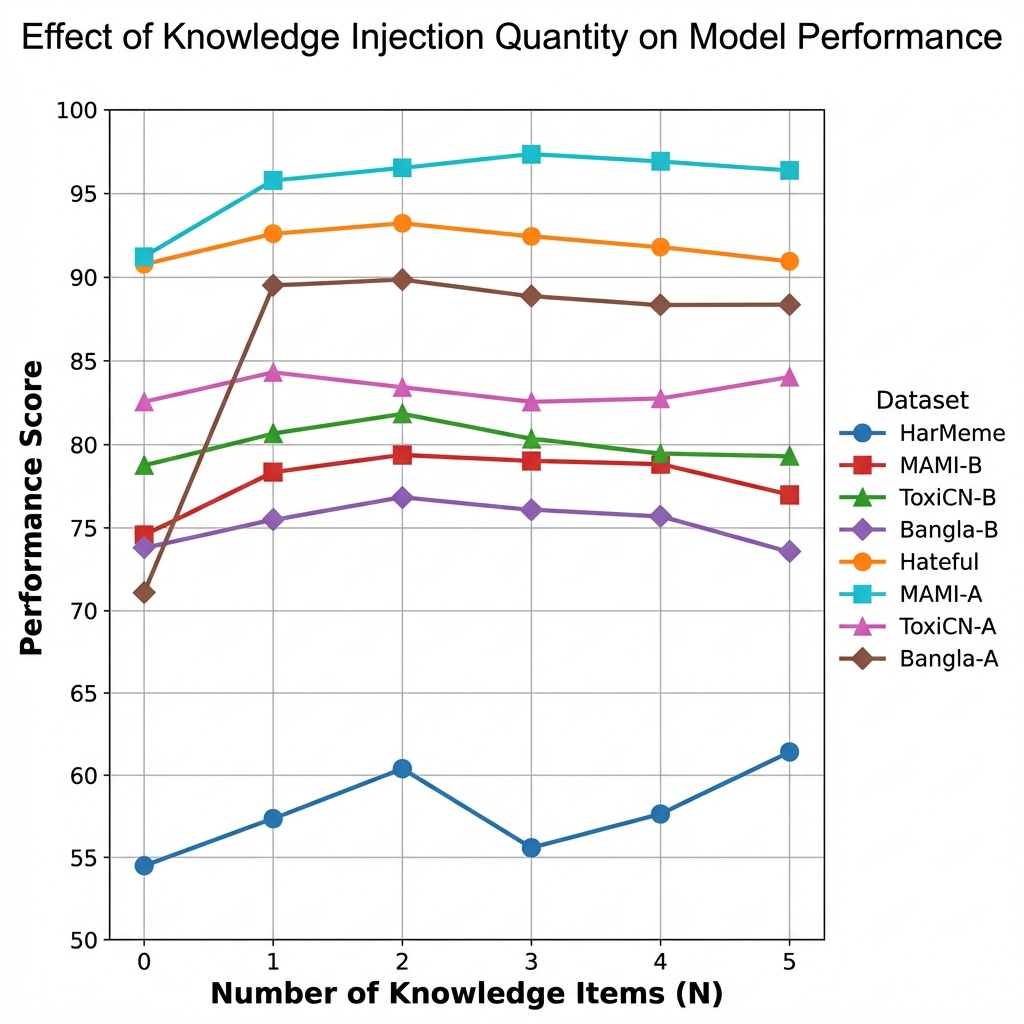}
\caption{Inverted-U relationship between knowledge injection quantity and model performance across eight datasets (measured by Macro-F1 for multi-class tasks and AUC/F1 for binary tasks). Most datasets achieve peak performance at $N=2$, demonstrating the optimal balance between information gain and noise suppression.}
\label{fig:knowledge_curve}
\end{figure}

A critical design choice in our framework is the number of knowledge items ($N$) injected per meme. We conduct comprehensive ablation studies varying $N$ from 0 to 5 to investigate the relationship between knowledge quantity and model performance.

The experimental results reveal a clear \textbf{inverted-U relationship} between knowledge injection quantity and model performance. Detailed results are shown in Table~\ref{tab:ablation_multiclass} and Table~\ref{tab:ablation_binary}, with the trend visualized in Figure~\ref{fig:knowledge_curve}.

\textbf{Low-N Phase ($0 \le N < 2$):} The model is in a ``knowledge-hungry'' state. Each additional piece of external knowledge significantly bridges the semantic gap between visual features and textual labels, providing critical discriminative cues.

\textbf{Optimal Balance Point ($N = 2$):} Across most datasets, $N = 2$ achieves peak or near-peak performance. This suggests that two core knowledge items are typically sufficient to cover the primary metaphorical entities and their logical conflict points within a meme.

\textbf{High-N Phase ($N > 2$):} As $N$ increases beyond the optimal point, performance begins to decline. Excessive knowledge introduces task-irrelevant redundant information, reducing the system's signal-to-noise ratio (SNR).

\textbf{Information-Theoretic Interpretation.} The inverted-U relationship can be formalized as: $I_{\text{eff}}(N) = I_{\text{useful}}(N) - \alpha \cdot I_{\text{noise}}(N)$, where $\alpha$ is a noise sensitivity coefficient. Our empirical finding that $N = 2$ is optimal provides guidance for practical deployment: knowledge augmentation should focus on quality over quantity.

\subsubsection{Effect of Dual-Head Architecture}

\FloatBarrier
\begin{table}[!htbp]
\centering
\caption{Ablation Study: From General LMM to KID.}
\label{tab:ablation_dualhead}
\resizebox{\columnwidth}{!}{
\begin{tabular}{lccc}
\toprule    
Configuration & HarMeme & MAMI B & ToxiCN\_MM B \\
\midrule
\multicolumn{4}{l}{\textit{General LMM (No Task-Specific Training)}} \\
Gemini-2.5-Flash (Zero-shot) & 35.41 & 73.44 & 40.63 \\
Gemini-2.5-Flash (Few-shot) & 53.56 & 75.34 & 56.32 \\
\midrule
\multicolumn{4}{l}{\textit{Task-Specific Fine-Tuning}} \\
SFT (Generation Head Only) & 60.84 & 70.35 & 74.23 \\
Dual-Head SFT & 63.40 (+2.56) & 75.36 (+5.01) & 78.29 (+4.06) \\
Dual-Head + Knowledge (KID) & \textbf{65.50} (+4.66) & \textbf{79.60} (+9.25) & \textbf{81.84} (+7.61) \\
\bottomrule
\end{tabular}
}
\par\smallskip\footnotesize\raggedright
\textit{Note:} All metrics reported as Macro-F1.
\end{table}

Table~\ref{tab:ablation_dualhead} demonstrates \textbf{the necessity of both heads in our dual-head architecture}. The ``SFT (Generation Only)'' configuration represents a standard LMM fine-tuning approach that relies solely on the language modeling head---the model outputs classification labels in natural language form without a dedicated discriminative classifier. In contrast, ``Dual-Head SFT'' adds an MLP-based classification head that directly maps the last-token representation to class logits.

The ablation results reveal that \textbf{both heads contribute uniquely and are essential}:
\begin{itemize}
    \item \textbf{The classification head} provides stable decision boundaries and consistent numerical outputs, contributing +2.56\% to +5.01\% Macro-F1 improvements across datasets. This head is critical for achieving competitive quantitative metrics (AUC, F1) required in content moderation systems.
    \item \textbf{The semantic generation head} acts as an auxiliary regularization task that constrains the model to produce outputs consistent with the natural language label space, thereby preventing overfitting to superficial visual-textual correlations. When removed (i.e., training only with a classification objective), preliminary experiments showed increased overfitting on smaller datasets.
\end{itemize}

When combined with knowledge injection, the dual-head architecture achieves even more pronounced improvements (+7.61\% on ToxiCN\_MM), demonstrating the synergistic effect between structured knowledge and the dual-head design. This confirms that the joint optimization of ``semantic label generation'' (generation head) and ``decision stability'' (classification head) is central to our framework's success.

\textbf{Why Entity-Anchored Knowledge Injection Works.} The success of our entity-anchored knowledge injection can be attributed to its addressing of the fundamental ``knowledge-context disconnect'' problem. Our structured $\langle$Entity$\rangle$-[Knowledge] format creates explicit bindings between visual elements and their cultural/contextual significance, informed by cognitive theories of meme comprehension.

\subsubsection{Effect of Knowledge Injection Format}

\begin{table}[t]
\centering
\caption{Ablation Study: Knowledge Injection Format.}
\label{tab:ablation_format}
\resizebox{\columnwidth}{!}{
\begin{tabular}{lcccc}
\toprule
Format & HarMeme & MAMI B & ToxiCN B & Bangla B \\
\midrule
Appended & 62.83/77.42 & 76.45/84.80 & 78.92/83.54 & 73.26/76.84 \\
Inline (Ours) & \textbf{65.50/80.70} & \textbf{79.60/87.50} & \textbf{81.84/86.42} & \textbf{76.82/80.18} \\
\midrule
$\Delta$ & +2.67/+3.28 & +3.15/+2.70 & +2.92/+2.88 & +3.56/+3.34 \\
\bottomrule
\end{tabular}
}
\par\smallskip\footnotesize\raggedright
\textit{Note:} ``Inline'' refers to our entity-anchored approach where knowledge immediately follows each entity; ``Appended'' concatenates all knowledge as a glossary after the description. All metrics reported as Macro-F1 / ACC.
\end{table}

To validate the importance of entity-anchored positioning, we compare two knowledge injection formats:

\begin{itemize}
    \item \textbf{Inline (Entity-Anchored)}: Knowledge is inserted immediately after each entity, forming $\langle$Entity$\rangle$[Knowledge] pairs within the description. For example: ``...mocking $\langle$Ghotis$\rangle$[a colloquial term for people from West Bengal...] ...with $\langle$Jio$\rangle$[a telecom provider...]...''
    
    \item \textbf{Appended (Glossary-Style)}: The description remains unmodified, with all knowledge concatenated as a separate glossary section at the end. For example: ``...mocking Ghotis...with Jio... [Knowledge] Ghotis: ... Jio: ...''
\end{itemize}

Table~\ref{tab:ablation_format} presents the comparison results across four multi-class datasets. The inline format consistently outperforms the appended format by 2.67--3.56 percentage points in Macro-F1 across all datasets.

\textbf{Analysis.} The performance gap can be attributed to three factors: (1) \textit{Reduced association burden}: In the inline format, the model directly receives entity-knowledge bindings without needing to establish cross-positional mappings; (2) \textit{Contextual activation}: Knowledge presented in-situ activates more effectively within the semantic context where the entity appears; (3) \textit{Attention preservation}: In long sequences, appended knowledge at the end may suffer from attention dilution, particularly in decoder-based models where earlier tokens receive more processing depth.

These findings empirically validate our entity-anchored design choice as a principled approach to structured knowledge injection.

\subsubsection{Zero-shot vs. Few-shot vs. Fine-tuning}

Table~\ref{tab:ablation_dualhead} presents a comprehensive comparison across different learning paradigms, revealing a clear progression from general LMM capabilities to task-specific knowledge-injected fine-tuning.

\textbf{The Knowledge Acquisition Gap.} Zero-shot Gemini-2.5-Flash, despite possessing extensive world knowledge, achieves only 35.41\% Macro-F1 on HarMeme---significantly below even simple SFT approaches (60.84\%). This gap illustrates the core challenge: \textit{general knowledge alone is insufficient without task-specific contextual grounding}. Few-shot learning partially bridges this gap (+18.15\% on HarMeme) by providing in-context examples, but the improvements remain limited because the model cannot generalize the implicit ``evidence-knowledge-label'' reasoning patterns from a handful of examples.

\textbf{The Role of Task-Specific Supervision.} Our framework addresses this through a two-stage knowledge integration strategy:
\begin{itemize}
    \item \textbf{Instance-level knowledge acquisition}: We leverage powerful LMMs (Gemini-2.5-Flash) to generate meme-specific contextual knowledge for each training sample, explicitly binding background knowledge to visual entities within individual memes.
    \item \textbf{Dataset-level supervised learning}: The fine-tuned model then learns the classification criteria across the entire training set, internalizing consistent ``evidence-knowledge-label'' reasoning patterns that generalize to unseen samples.
\end{itemize}

This two-stage approach---where knowledge is acquired at the individual meme level but classification patterns are learned at the dataset level---explains why KID (+4.66\% over SFT on HarMeme) substantially outperforms both in-context learning and vanilla fine-tuning approaches.

\FloatBarrier
\section{Case Study}
\label{sec:casestudy}

\begin{figure*}[!t]
\centering
\includegraphics[width=\textwidth]{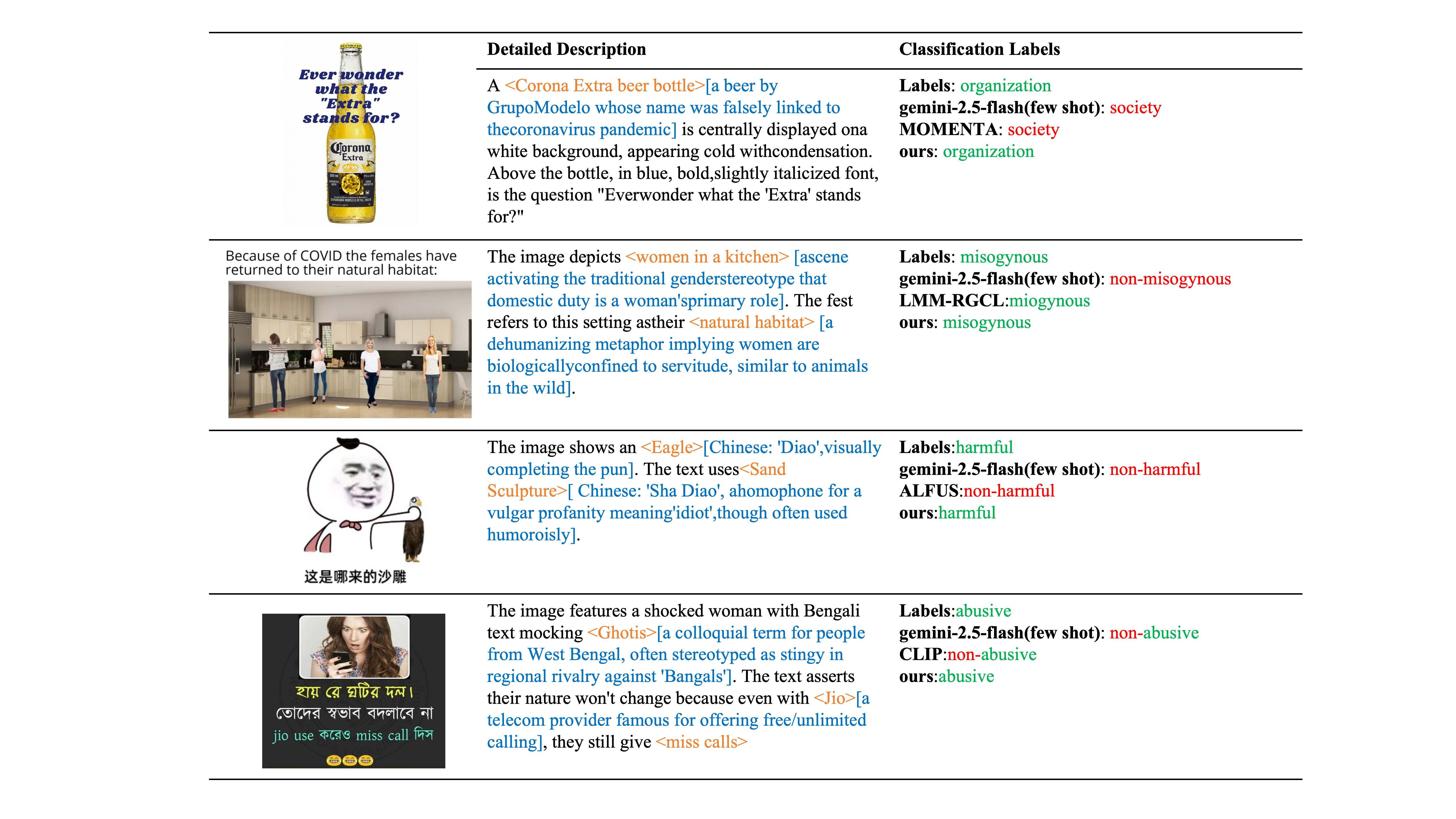}
\caption{Case Study Examples: Four samples illustrating entity ambiguity (Corona beer brand-pandemic association), implicit stereotype (kitchen-gender role confinement), cultural context (Chinese internet slang disambiguation), and low-resource language generalization (Bengali regional discrimination).}
\label{fig:casestudy}
\end{figure*}

To demonstrate the robustness of KID in complex semantic recognition, we conduct a qualitative analysis on four representative samples (Figure~\ref{fig:casestudy}). These samples correspond to four distinct multimodal reasoning challenges: entity ambiguity, implicit stereotypes, cultural context disambiguation, and low-resource language generalization.

\subsection{Entity Ambiguity Resolution}

The first case shows a conflict between visual and textual modalities. The image presents a commercial product (Corona Extra beer), while the text triggers associations with the COVID-19 pandemic through the ambiguous term ``Corona.'' Such conflicts often cause traditional models to rely too heavily on text, where high-frequency statistical correlations overshadow visual evidence. Indeed, the baseline model (Gemini-2.5-Flash under few-shot prompting) misclassifies this sample as targeting ``Society,'' imagining a pandemic-related societal context rather than recognizing the brand-level attack.

Our KID framework resolves this conflict through entity-anchored knowledge injection. By explicitly providing the definition ``[a beer by Grupo Modelo...],'' we introduce structured reasoning guidance that grounds the model's reasoning in the visual entity rather than textual associations. The semantic generation head articulates this correction, enabling the classification head to correctly identify the target as ``Organization.''

\subsection{Implicit Stereotype Recognition}

The second case shows how image and text work together to create toxicity. Neither the visual modality (smiling women in a kitchen) nor the textual modality (``natural habitat'') is overtly toxic alone. The toxicity emerges from their interaction---a metaphorical mapping that reinforces gender stereotypes by implying that the kitchen is women's ``natural'' domain. Existing models often struggle with such cases because they treat modalities independently, failing to bridge the gap between benign surface features and deeper ideological meanings. The baseline model predicts ``Non-misogynous,'' seemingly misled by the positive visual sentiment.

KID successfully captures this synergy through structured knowledge injection. The injected context ``[...activates stereotype... 'natural habitat' is a dehumanizing metaphor]'' serves as a semantic bridge, explicitly linking the visual scene to the abstract concept of role confinement. This enables the model to interpret the ``kitchen'' not merely as a physical location, but as a symbol of gender-based oppression, correctly identifying the misogynous intent.

\subsection{Cultural Context Disambiguation}

The third case presents a semantic shift challenge rooted in cultural linguistics. The Chinese text contains the term ``sha diao'' (literally ``Sand Sculpture''), a homophonic internet slang that can function either as self-deprecating humor or as a vulgar insult depending on context. The pointing gesture in the image reframes this ambiguous term from general humor to targeted attack. The baseline model predicts ``Non-harmful,'' defaulting to the sanitized, humorous interpretation while ignoring the visual cue that signals aggressive intent.

Our framework handles this nuance through the complementary dual-head architecture. The semantic generation head acknowledges the term's dual nature and cultural context, while the classification head---trained on the toxic dataset---detects the aggressive combination of gesture and slang. The model correctly identifies this as a harmful attack, demonstrating its capability in handling pragmatic ambiguity across cultural contexts.

\subsection{Low-Resource Language Generalization}

The fourth case demonstrates our framework's capability in handling low-resource languages, specifically Bengali. The meme features a shocked woman with Bengali text mocking ``Ghotis''---a colloquial term for people from West Bengal, often stereotyped as stingy in regional rivalry against Bengalis from Bangladesh. The text asserts that ``their nature won't change'' because even with Jio (a telecom provider famous for offering free/unlimited calling), they still give missed calls to avoid spending money.

Baseline models struggle with such content for two reasons: (1) limited Bengali training data in pre-training corpora leads to poor semantic understanding of regional slurs, and (2) the absence of cultural-pragmatic knowledge about inter-regional stereotypes between West Bengal and Bangladesh. Both Gemini-2.5-Flash (few-shot) and ALFUS predict ``Non-abusive,'' failing to detect the subtle regional discrimination embedded in the colloquial expression.

KID addresses this through entity-anchored knowledge injection that explicitly bridges the linguistic and cultural gap. The injected context provides cultural annotation explaining that ``Ghotis'' is a derogatory term used in regional rivalry, and that the ``miss call'' reference reinforces negative stereotypes about stinginess. This explicit contextualization enables our dual-head model to correctly classify this as abusive content, demonstrating the framework's robust cross-lingual transfer capability---consistent with our quantitative findings on BanglaAbuseMeme (+19.44 percentage points over baselines).

\section{Conclusion}
\label{sec:conclusion}

In this paper, we presented KID, a Knowledge-Injected Dual-Head Learning framework for harmful meme detection. Our approach addresses the critical challenge of ``having knowledge but not knowing how to use it'' that plagues existing multimodal models when processing culturally-laden harmful memes.

The key contributions of our work include: (1) an entity-anchored knowledge construction paradigm that creates structured chains, explicitly binding external commonsense to visual entities within memes; (2) a dual-head architecture that jointly optimizes label generation and classification through multi-task learning, providing regularization effects that prevent overfitting; and (3) empirical demonstration that optimal knowledge injection quantity ($N = 2$) achieves the best balance between information gain and noise suppression, and that entity-anchored inline injection consistently outperforms appended glossary-style formats by 2.67--3.56 percentage points.

Extensive experiments across five multilingual datasets---spanning English, Chinese, and low-resource Bengali---validate the effectiveness of our framework. KID achieves leading performance on both binary and multi-class harmful meme detection tasks, with particularly notable improvements on challenging benchmarks such as Hateful Memes (+2.14 percentage points AUC) and MAMI (+6.0 percentage points AUC), as well as exceptional cross-lingual transfer to low-resource Bengali (+19.44 percentage points Macro-F1).

\textbf{Limitations.} Despite the strong performance, our framework has several limitations. First, the quality of injected knowledge depends on the teacher LMM's world knowledge and generation capabilities; for extremely niche or emerging cultural phenomena, the teacher may generate incomplete or inaccurate knowledge. Second, the test-time knowledge injection step requires an additional LMM API call, increasing inference latency. Third, the optimal knowledge injection strategy for multi-label scenarios may differ from single-label settings and warrants further investigation.

\textbf{Future Work.} We plan to explore more efficient knowledge injection mechanisms that reduce inference latency, investigate adaptive strategies for determining optimal knowledge quantity per sample, and extend the framework to emerging forms of multimodal harmful content such as short-form videos.

\section*{Declaration of generative AI and AI-assisted technologies in the manuscript preparation process}
During the preparation of this work, the authors used ChatGPT to assist with language polishing and improving readability. After using this tool, the authors reviewed and edited the content as needed and take full responsibility for the content of the published article.

\section*{CRediT authorship contribution statement}

\textbf{Yaocong Li}: Methodology, Investigation, Formal analysis, Visualization, Writing -- Original Draft, Writing -- Review \& Editing.
\textbf{Le Zhang}: Methodology, Supervision (experiments), Writing -- Review \& Editing, Funding acquisition.
\textbf{Leihan Zhang}: Conceptualization, Methodology (guidance), Formal analysis (discussion), Writing -- Review \& Editing, Project administration, Funding acquisition.
\textbf{Qiang Yan}: Project administration, Writing -- Review \& Editing.

\section*{Declaration of competing interest}
The authors declare that they have no known competing financial interests or personal relationships that could have appeared to influence the work reported in this paper.

\section*{Acknowledgement}
This research was funded by the National Natural Science Foundation of China (NSFC) under Grant Nos. 62102044 and 72501034.
\bibliographystyle{elsarticle-num}
\bibliography{KID}

\end{document}